%% file: egpaper_arxiv.tex
\documentclass[final]{iccv}

\usepackage[accsupp]{axessibility}
\usepackage{times}
\usepackage{epsfig}
\usepackage{graphicx}
\usepackage{amsmath}
\usepackage{amssymb}
\usepackage[utf8]{inputenc}
\usepackage{times}
\usepackage{epsfig}
\usepackage[version=3]{mhchem}
\usepackage{graphicx}
\usepackage{comment}
\usepackage{ifthen}
\usepackage{float}
\usepackage{xspace}
\usepackage[dvipsnames,table,xcdraw]{xcolor}
\usepackage{amsfonts}
\usepackage{amsmath}
\usepackage{booktabs} 
\usepackage{nicefrac} 
\usepackage{multirow}
\usepackage{amssymb}
\usepackage{enumitem} % for itemized bullets
\usepackage{euscript} % for curly N
\usepackage{dsfont}
\usepackage{placeins}
\usepackage{algorithm}
\usepackage{listings}
\usepackage{subcaption}
\usepackage{makecell}
\usepackage{dsfont}
\usepackage{bbm}
\usepackage{hhline}
\usepackage{cellspace}
\usepackage{mathtools}

\usepackage[pagebackref=true,breaklinks=true,letterpaper=true,colorlinks,bookmarks=false]{hyperref}


\newcommand{\Lagr}{\mathcal{L}}

\def\cvprPaperID{7171} % *** Enter the CVPR/ICCV Paper ID here
\def\confName{ICCV}
\def\confYear{2023}

\begin{document}

%%%%%%%%% TITLE
\title{2D-3D Interlaced Transformer for\\Point Cloud Segmentation with Scene-Level Supervision}

\author{
Cheng-Kun Yang$^{1,2}$ \quad
Min-Hung Chen$^3$ \quad
Yung-Yu Chuang$^1$ \quad 
Yen-Yu Lin$^{4,5}$
\\
$^1$National Taiwan University \quad 
$^2$MediaTek \quad 
$^3$NVIDIA \quad  \\
$^4$National Yang Ming Chiao Tung University \quad
$^5$Academia Sinica}

\maketitle

\input{0_abstract}

\input{1_introduction}

\input{2_related-work}

\input{3_method}

\input{4_experiment}

\input{5_conclusion}

\input{6_supplementary}

% \clearpage
{\small
\bibliographystyle{ieee_fullname}
\bibliography{egbib}
}

\end{document}

%% file: 0_abstract.tex
\begin{abstract}
We present a \textbf{M}ultimodal \textbf{I}nterlaced \textbf{T}ransformer (\textbf{MIT}) that jointly considers 2D and 3D data for weakly supervised point cloud segmentation.
Research studies have shown that 2D and 3D features are complementary for point cloud segmentation. 
However, existing methods require extra 2D annotations to achieve 2D-3D information fusion.
Considering the high annotation cost of point clouds, effective 2D and 3D feature fusion based on weakly supervised learning is in great demand.
To this end, we propose a transformer model with two encoders and one decoder for weakly supervised point cloud segmentation using only scene-level class tags.
Specifically, the two encoders compute the self-attended features for 3D point clouds and 2D multi-view images, respectively.
The decoder implements interlaced 2D-3D cross-attention and carries out implicit 2D and 3D feature fusion.
We alternately switch the roles of queries and key-value pairs in the decoder layers.
It turns out that the 2D and 3D features are iteratively enriched by each other.
Experiments show that it performs favorably against existing weakly supervised point cloud segmentation methods by a large margin on the S3DIS and ScanNet benchmarks. 
The project page will be available at  \small\url{https://jimmy15923.github.io/mit_web/}.
\end{abstract}

%% file: 1_introduction.tex
\vspace{-0.1in}
\section{Introduction} \label{sec:intro}

Point cloud segmentation offers rich geometric and semantic information of a 3D scene, thereby being essential to many 3D applications, such as scene understanding~\cite{behley2019semantickitti, choy20194d, hou2021exploring, landrieu2018large, qi2020imvotenet}, augmented reality~\cite{alexiou2017towards, park2008multiple}, and autonomous driving~\cite{chen20203d, chen2017multi, geiger2013vision}.
However, developing reliable models is time-consuming and challenging due to the need for vast per-point annotations and the difficulty in capturing detailed semantic clues from textureless point clouds.

\input{image/teaser}

Research efforts have been made to address the aforementioned issues.
Several methods are proposed to derive point cloud segmentation models using various weak supervisions, such as sparsely labeled points~\cite{kweon2022joint,liu2021one,xu2020weakly,zhang2021weakly}, bounding box labels~\cite{chibane2021box2mask}, subcloud-level annotations~\cite{wei2020multi}, and scene-level tags~\cite{ren20213d, yang2022mil}. 
These weak annotations are cost-efficient and can significantly reduce the annotation burden.
On the other hand, recent studies~\cite{hu2021bidirectional,jaritz2019multi,kundu2020virtual,liu2022petr,robert2022learning,wang2022semaffinet,wang2022bridged,yang2019learning} witness the remarkable success of 2D vision, and utilize 2D image features to enhance the 3D recognition task. 
They show promising results because 2D detailed texture clues are well complementary to 3D geometry features.

Although 2D-3D fusion is effective, current methods require extra annotation costs for 2D images.
To the best of our knowledge, no prior work has explored fusing 2D-3D features under \textit{extremely weak supervision}, where only scene-level class tags of the 3D scene are given.
It is challenging to derive a segmentation model that leverages both 2D and 3D data under scene-level supervision, as no per-point/pixel annotations or per-image class tags are available to guide the learning process.
Furthermore, existing 2D-3D fusion methods require camera poses or depth maps to establish pixel-to-point correspondences, adding extra burdens on data collection and processing.
In this work, we address these difficulties by proposing a  \textbf{M}ultimodal \textbf{I}nterlaced \textbf{T}ransformer (\textbf{MIT}) that works with scene-level supervision and can implicitly fuse 2D, and 3D features without camera poses and depth maps.

Our  MIT is a transformer model with two encoders and one decoder, and can carry out weakly supervised point cloud segmentation.
As shown in Figure~\ref{fig:teaser}, the input to our method includes the 3D point cloud, multi-view images, and scene-level tags of a scene.
The two encoders utilize the self-attention mechanism to extract the features of the 3D point cloud and the 2D multi-view images, respectively.
The decoder computes the proposed interlaced 2D-3D attention and can implicitly fuse the 2D and 3D data.

Specifically, one encoder is derived for 3D feature extraction, where the voxels of the input point cloud yield the data tokens.
The other encoder is for 2D multi-view images, where images serve as data tokens.
Also, the multi-class tokens~\cite{xu2022multi} are included to match the class-level annotations. 
The encoders capture long-range dependencies and aggregate class-specific features for their respective modalities.

The decoder comprises 2D-3D interlaced layers, and is developed to fuse 2D and 3D features, where the correspondences between 3D voxels and 2D views are implicitly computed via cross-attention.
In odd layers of the decoder, 3D voxels are enriched by 2D image features, while in even layers, 2D views are augmented by 3D geometric features.
Specifically, in each odd layer, each 3D voxel serves as a {\em query}, while 2D views act as {\em key-value pairs}. 
Through cross-attention, a query is a weighted combination of the values. 
Together with residual learning, this query (3D voxel) is characterized by the fused 3D and 2D features.
In each even layer, the roles of 3D voxels and 2D views switch: 3D voxels and 2D views become key-value pairs and queries, respectively. 
This way, 2D views are described by the augmented 2D and 3D features.

By leveraging multi-view information without extra annotation effort, our proposed MIT effectively fuses the 2D and 3D features and significantly improves 3D point cloud segmentation.
The main contribution of this work is threefold. 
First, to the best of our knowledge, we make the first attempt to fuse 2D-3D information for point cloud segmentation under scene-level supervision.
Second, we enable this new task by presenting a new model named Multimodal Interlanced Transformer (MIT) that implicitly fuses 2D-3D information via interlaced attention, which does not rely on camera pose information.
Besides, a contrastive loss is developed to align the class tokens across modalities
Third, our method performs favorably against existing methods on the large-scale ScanNet~\cite{dai2017scannet} and S3DIS~\cite{armeni20163d} benchmarks.

%% file: image/teaser.tex
\begin{figure}[t]
    \centering
    \includegraphics[width=.95\linewidth]{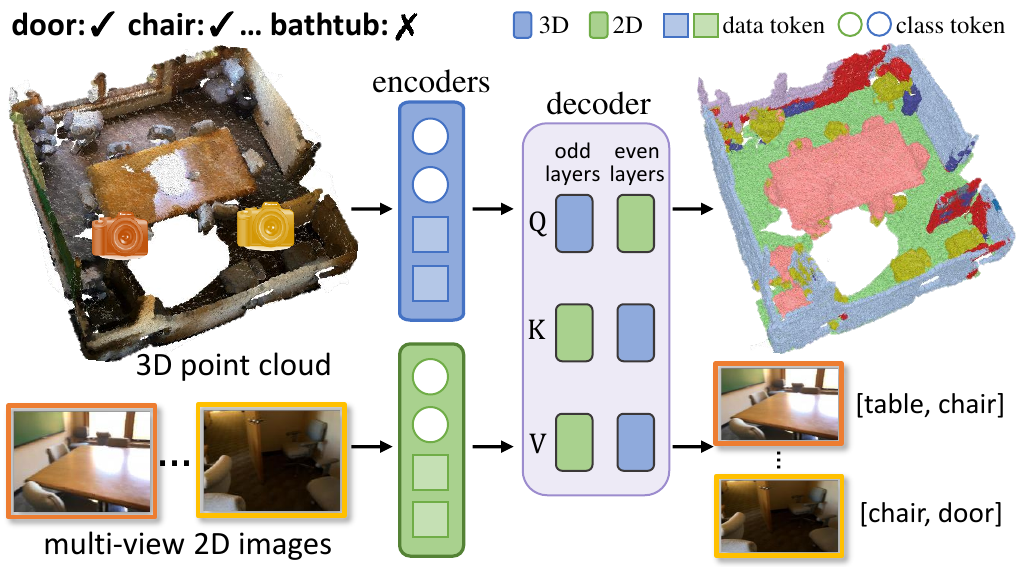}
    \vspace{-0.05in}
    \caption{
    Overview of the \textbf{M}ultimodal \textbf{I}nterlaced \textbf{T}ransformer (\textbf{MIT}).
    The input includes a 3D point cloud, multi-view 2D images, and class-level tags of a scene.
    Our method is a transformer model with two encoders and one decoder.
    The two encoders compute features for 3D voxel tokens and 2D view tokens, respectively.
    The decoder conducts interlaced 2D-3D attention and carries out 2D and 3D feature fusion.
    In its odd layers, 3D voxels serve as queries and are enriched by the semantic features of 2D views, acting as key-value pairs. 
    In the even layers, the roles of 3D voxels and 2D views switch: 2D views are described by additional 3D geometric features.
    }
    \label{fig:teaser}
\vspace{-0.10in}
\end{figure}

%% file: 2_related-work.tex
\section{Related Work}

\paragraph{Weakly supervised point cloud segmentation.}
This task aims at learning a point cloud segmentation model using weakly annotated data, such as sparsely labeled points~\cite{hou2021exploring,hu2022sqn,liu2021one,li2022hybridcr,xu2020weakly,tao2022seggroup,wu2022dual,sun2022image,zhang2021perturbed,zhang2021weakly,shi2022weakly,unal2022scribble,liu2022weaklabel3d,lee2023gaia}, box-level labels~\cite{chibane2021box2mask}, subcloud-level labels~\cite{wei2020multi,yang2022mil} and scene-level labels~\cite{ren20213d, kweon2022joint}.
Significant progress has been made in the setting of using sparsely labeled points: The state-of-the-art methods~\cite{liu2021one,yu2022data,hu2022sqn} show comparable performances with supervised ones. 
These methods usually utilize self-supervised pre-training~\cite{hou2021exploring,liu2021one}, graph propagation~\cite{tao2022seggroup,liu2021one}, and contrastive learning~\cite{li2022hybridcr,zhang2021perturbed} to derive the models. 
Despite the effectiveness, they require at least one annotated point for each category in a scene. 
Hence, it is not straightforward to extend these methods to work with scene-level or subcloud-level annotations. 

In this work, we aim to develop a segmentation method based on a more challenging setting of using scene-level annotations. The literature about point cloud segmentation with scene-level annotations is relatively rare.
Yang~\etal~\cite{yang2022mil} derive a transformer by applying multiple instance learning to paired point clouds. However, their performance is much inferior to fully supervised methods.
Kweon and Yoon~\cite{kweon2022joint} leverage 2D and 3D data for point cloud segmentation by introducing additional image-level class tags, which require extra annotation efforts.
Our method compensates for the lack of point-level or pixel-level annotations by integrating additional 2D features while using scene-level annotation only.

\input{image/model}

\vspace{-0.15in}
\paragraph{2D and 3D fusion for point cloud applications.}
Given the accessible or syntheticable~\cite{kundu2020virtual} 2D images in most 3D dataset, 
research studies~\cite{genova2021learning,hou2021pri3d,hu2021bidirectional,jaritz2019multi,kundu2020virtual,kweon2022joint,li2020attention,liu2022petr,robert2022learning,wang2022semaffinet,wang2022bridged,yan20222dpass,yang2019learning,yuan2022x,Sun_2020_ACCV,wu3dsegmenter,yang2022towards} explore 2D data to enhance 3D applications.
Hu~\etal~\cite{hu2021bidirectional} and Robert~\etal~\cite{robert2022learning} construct a pixel-point mapping matrix to fuse 2D and 3D features for point cloud segmentation. 
Despite the effectiveness, existing methods rely on camera poses and/or depth maps to build the correspondences between the 2D and 3D domains.
In contrast, our method learns a transformer with interlaced 2D-3D attention, enabling the implicit integration of 2D and 3D features without the need for camera poses or depth maps.

\vspace{-0.15in}
\paragraph{Query and key-value pair swapping.}
Cross-attention is widely used in the transformer decoder. 
It captures the dependency between queries and key-value pairs. 
Umam~\etal~\cite{umam2022point} and Lim~\etal~\cite{kim2021setvae} swap queries and key-value pairs for point cloud decomposition and generation, respectively. 
Different from their methods working with data in a domain, our method generalizes query and key-value pair swapping to cross-domain feature fusion.
In addition, we develop a contrastive loss for 2D and 3D feature alignment.

%% file: image/model.tex
\begin{figure*}[t]
    \centering
    \includegraphics[width=.95\linewidth]{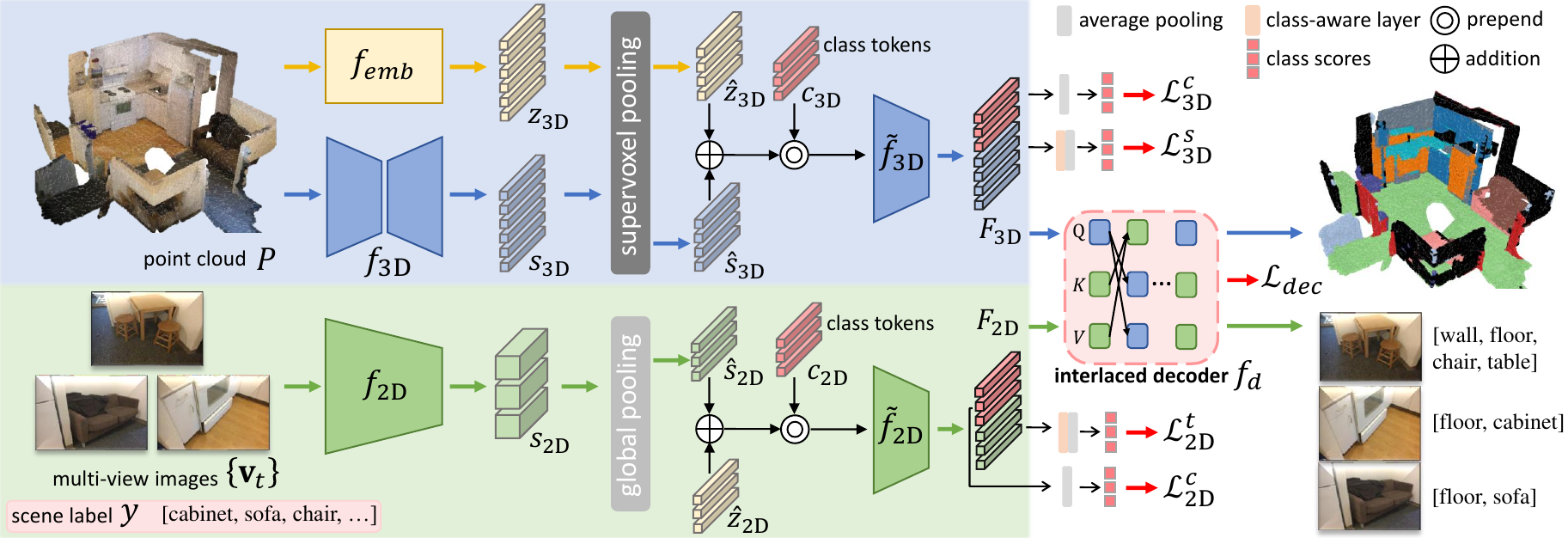}
    \vspace{-0.05in}
    \caption{
    An overview of our \textbf{M}ultimodal \textbf{I}nterlaced \textbf{T}ransformer (\textbf{MIT}) for weakly supervised point cloud segmentation.
    It is a transformer-based model with two encoders,  $\tilde{f}_{\text{3D}}$ and $\tilde{f}_{\text{2D}}$, for modality-specific feature extraction and one decoder, $f_d$, for feature fusion.
    The 2D and 3D pooled features, $\hat{s}_\text{2D}$ and $\hat{s}_\text{3D}$, are added to each learnable position embedding ($\hat{z}_\text{2D}$ and $\hat{z}_\text{3D}$), and further prepended with the class tokens and passed through the encoders to obtain self-attended features, $F_{\text{2D}}$ and $F_{\text{3D}}$.
    The predicted class scores for each modality are obtained through average pooling and class-aware layers.
    }
    
    \label{fig:model}
    \vspace{-0.05in}
\end{figure*}

%% file: 3_method.tex
\section{Proposed Method}

We present the proposed method in this section. 
We first give the problem statement in Section~\ref{sec:problem}. 
Then, we specify the developed MIT with an encoder-decoder architecture in Section~\ref{sec:encoder} and Section~\ref{sec:decoder}. 
Finally, the implementation details are provided in Section~\ref{sec:implementation}. 

\subsection{Problem Statement}
\label{sec:problem}

We are given a set of $N$ point clouds as well as their corresponding RGB multi-view images and the class tag annotations, \ie, $\{P_n,V_n,\mathbf{y}_n\}_{n=1}^N$, where $P_n$ denotes the $n$th point cloud, $V_n$ represents the multi-view images, and $\mathbf{y}_n$ is the class-level labels.
Note that $P_n$, $V_n$, and $\mathbf{y}_n$ are acquired from the same scene.
Without loss of generality, we assume that each point cloud consists of $M$ points, \ie, $P_n = \{\mathbf{p}_{nm}\}_{m=1}^M$, where each point $\mathbf{p}_{nm} \in \mathbb{R}^6$ is represented by its 3D coordinate and RGB color.
The RGB multi-view images are grabbed in the same scene as $P_n$, and consist of a set of $T$ images, \ie, $V_n = \{\mathbf{v}_{nt}\}_{t=1}^T$.
Each image $\mathbf{v}_{nt} \in \mathbb{R}^{H \times W \times 3}$ is of resolution $H \times W$ with RGB channels. 
The class tags of $P_n$, \ie, $\mathbf{y}_n \in \mathbb\{0,1\}^{C}$, are a $C$-dimensional binary vector storing which categories are present, where $C$ is the number of categories of interest.

With the weakly annotated dataset $\{P_n,V_n,\mathbf{y}_n\}_{n=1}^N$, we aim to derive a model for point cloud segmentation that classifies each point of a testing cloud into one of the $C$ categories. 
Note that in this weakly supervised setting, neither points nor pixels are labeled, and camera poses are unavailable, making it challenging to enhance 3D point cloud segmentation by incorporating additional 2D features due to the absence of point/pixel supervision and explicit correspondences between 2D pixels and 3D points.
Furthermore, as multi-view images share the same scene-level class tag, the lack of individual class tag annotation for each view image may lead to an inaccurate semantic understanding of each image.

\vspace{-0.15in}
\paragraph{Method overview.}
Figure~\ref{fig:model} illustrates the network architecture of MIT, which comprises two transformer encoders, $\tilde{f}_\text{3D}$ and $\tilde{f}_\text{2D}$, and one decoder $f_d$.
The two encoders extract features for 3D point clouds and 2D multi-view images, respectively.
The decoder is developed for 2D-3D feature fusion, which utilizes cross-attention to connect 2D and 3D data implicitly.
They are elaborated in the following.

\subsection{Transformer Encoders}

\paragraph{3D point cloud feature extraction.}
A 3D backbone $f_\text{3D}$, \eg, MinkowskiNet~\cite{choy20194d} or PointNet++~\cite{qi2017pointnet++}, is applied to extract the point embedding $s_\text{3D} \in \mathbb{R}^{M \times D}$ for all $M$ points of a point cloud $P$.
Like WYPR~\cite{liu2021weakly}, we perform supervoxel partition using an unsupervised off-the-shelf algorithm~\cite{lafarge2012creating}. 
The 3D coordinates of $P$ are fed into a coordinate embedding module $f_{emb}$, which is composed of two $1 \times 1$ convolution layers with ReLU activation, to get the positional embedding $z_\text{3D} \in \mathbb{R}^{M \times D}$, where $D$ is the embedding dimension.
We aggregate both the point features and point positional embedding through supervoxel average pooling~\cite{liu2021one}, producing the supervoxel features $\hat{s}_\text{3D} \in \mathbb{R}^{S \times D}$ and pooled positional embedding $\hat{z}_\text{3D} \in \mathbb{R}^{S \times D}$, where $S$ is the number of the supervoxels in $P$. 
The supervoxel features are added to the positional embedding.

To learn the class-specific representation for fitting the scene-level supervision, we prepend $C$ learnable class tokens~\cite{xu2022multi} $c_\text{3D} \in \mathbb{R}^{C \times D}$ with $S$ supervoxel tokens.
Total $(C+S)$ tokens are fed into the transformer encoder $\tilde{f}_\text{3D}$. 
Through the self-attention mechanism, the dependencies of the class and supervoxel tokens are captured, producing the self-attended 3D features $F_\text{3D} \in \mathbb{R}^{(C + S) \times D}$.

\input{image/decoder}
\paragraph{2D multi-view images feature extraction.}
A 2D backbone network $f_\text{2D}$, \eg ResNet~\cite{he2016deep}, is employed to extract image features $s_\text{2D} \in \mathbb{R}^{T \times H^\prime \times W^\prime \times D}$, where $H^\prime = H/32$ and $W^\prime = W/32$.
We apply global average pooling to image features $s_\text{2D}$ along the spatial dimensions. 
The pooled image features $\hat{s}_\text{2D} \in \mathbb{R}^{T \times D}$ are added to the learnable positional embedding $\hat{z}_\text{2D} \in \mathbb{R}^{T \times D}$, producing $T$ view tokens.
Analogous to 3D feature extraction, another transformer encoder $\tilde{f}_\text{2D}$ is applied to $C$ class tokens $c_\text{2D} \in \mathbb{R}^{C \times D}$ and $T$ view tokens, obtaining the self-attended 2D features $F_\text{2D} \in \mathbb{R}^{(C + T) \times D}$.

\vspace{-0.15in}
\paragraph{Encoder optimization.}
\label{sec:encoder}
During training, we consider a point cloud $P$ and its associated $T$ multi-view images $\{\mathbf{v}_t\}$ and scene-level label $\mathbf{y}$.
The 2D and 3D self-attended features $F_\text{2D}$ and $F_\text{3D}$ are compiled as specified above.
We conduct the multi-label classification losses~\cite{ren20213d,xu2022multi} for optimization. 

For 3D attended features $F_\text{3D} \in \mathbb{R}^{(C + S) \times D}$, we divide it into $C$ class tokens $F_\text{3D}^{c} \in \mathbb{R}^{C \times D}$ and $S$ supervoxel tokens $F_\text{3D}^{s} \in \mathbb{R}^{S \times D}$.
For the class tokens $F_\text{3D}^{c}$, the $C$ class scores are estimated by applying average pooling along the feature dimension. 
The multi-label classification loss $\Lagr^c_\text{3D}$ is computed based on the estimated class scores and the scene-level ground-truth labels $\mathbf{y}$.
For the supervoxel tokens $F_\text{3D}^{s}$, we introduce a class-aware layer~\cite{sun2020mining}, \ie, a $1 \times 1$ convolution layer with $C$ filters, which maps the supervoxel tokens $F_\text{3D}^{s}$ into the class activation maps (CAM) $\tilde{F}_\text{3D}^{s} \in \mathbb{R}^{S \times C}$.
The estimated class scores are obtained by applying global average pooling to $\tilde{F}_\text{3D}^{s}$ along the dimension of supervoxels. 
The multi-label classification loss $\Lagr^s_\text{3D}$ is computed based on the class scores and label $\mathbf{y}$.
The loss for the 3D modality is defined by $\Lagr_\text{3D} = \Lagr^c_\text{3D}+\Lagr^s_\text{3D}$. 
For the self-attended 2D features $F_\text{2D} \in \mathbb{R}^{(C + T) \times D}$ of the $C$ class tokens and $T$ view tokens, the 2D loss is similarly defined by $\Lagr_\text{2D} = \Lagr^c_\text{2D}+\Lagr^t_\text{2D}$.
In sum, both encoders are derived in a weakly-supervised manner using the objective function 
\begin{equation}
\Lagr_{enc} = \Lagr_\text{2D} + \Lagr_\text{3D}.
\label{eq:enc_loss}
\end{equation} 
 
\subsection{Transformer Decoder}
\label{sec:decoder}

The two encoders produce self-attended 3D features $F_\text{3D}$ of $C+S$ tokens and 2D features $F_\text{2D}$ of $C+T$ tokens, respectively.
We propose a decoder that performs interlaced 2D-3D cross-attention for feature fusion.
The decoder $f_d$ in Figure~\ref{fig:model} is a stack of $R$ interlaced blocks.
Each interlaced block is composed of two successive decoder layers, as shown in Figure~\ref{fig:decoder}.
In the first layer of this block, 3D tokens are enriched by 2D features, while in the second layer, 2D tokens are enriched by 3D features. 

In the odd/first layer (the blue-shaded region in Figure~\ref{fig:decoder}), the $C+S$ tokens in $F_\text{3D}$ serve as the queries, while the $C+T$ tokens in $F_\text{3D}$ act as the key-value pairs.
Through scaled dot-product attention~\cite{vaswani2017attention}, the cross-modal attention matrix $A \in \mathbb{R}^{(C+S) \times (C+T)}$ (the yellow-shaded region) is computed to store the consensus between the 3D tokens and 2D tokens.
As we focus on exploring the relationships between 3D tokens and merely 2D view tokens in this layer, we ignore the attention values related to the 2D class tokens in $A$.
Specifically, only the query-to-view attention values $A_{q2v} \in \mathbb{R}^{(C+S) \times T}$ (green dots in Figure~\ref{fig:decoder}) are considered.
This is implemented by applying submatrix extraction to the attention matrix $A$ and the value matrix $V \in \mathbb{R}^{(C+T) \times D}$, \ie, $A_{q2v}=A[1 \colon C+S, C+1 \colon C+T]$  and $V_d=V[C+1 \colon C+T, \colon]$.

After applying the softmax operation to $A_{q2v}$, we perform matrix multiplication between the query-to-view attention matrix $A_{q2v}$ and the masked value matrix $V_d$. 
This way, each query (3D token) is a weighted sum of the values (2D view tokens).
Together with a residual connection, the resultant 3D tokens $F_{\hat{\text{3D}}}$ are enriched by 2D features.
It turns out that implicit feature fusion from 3D features to 2D features is carried out without using annotated data.

In the even/second layer (the green-shaded region in Figure~\ref{fig:decoder}), the roles of $F_{\hat{\text{3D}}}$ and $F_{\text{2D}}$ switch: 
The former serves as the key-value pairs while the latter yields the queries.
After a similar procedure, the resultant 2D tokens $F_{\hat{\text{2D}}} \in \mathbb{R}^{(C+T) \times D}$ are augmented with 3D information.
$F_{\hat{\text{2D}}}$ and  $F_{\hat{\text{3D}}}$ are the output of the interlaced block. 
By stacking $R$ interlaced blocks, the proposed decoder is built to fuse 2D and 3D features iteratively.

\vspace{-0.15in}
\paragraph{Decoder optimization.}
In the last interlaced block, the 2D class scores and 3D class scores can be estimated by applying average pooling to the corresponding class tokens.
The multi-label classification losses for 2D $\Lagr_{\tilde{\text{2D}}}$ and 3D $\Lagr_{\tilde{\text{3D}}}$ data can be computed between the ground truth and the estimated class scores.

To mine additional supervisory signals, we employ contrastive learning on the class-to-class attention matrix $A_{c2c} = A[1\colon C, 1\colon C] \in \mathbb{R}^{C \times C}$. 
Though the 2D class tokens and 3D class tokens attend to respective modalities, they share the same class tags. 
Hence, the attention value between a pair of class tokens belonging to the same class should be larger than those between tokens of different classes, which can be enforced by the N-pair loss~\cite{radford2021learning}.  
We employ this regularization in all attention matrices in the decoder layers
\begin{equation}
\begin{aligned}
\label{eq:npair}
    \Lagr_{con} =  \frac{1}{2R} \sum_{r=1}^{2R} \sum_{i=1}^C -\log \frac{A^r_{ii}}{\sum_{j=1}^C A^r_{ij}}  \\
+    \frac{1}{2R} \sum_{r=1}^{2R} \sum_{j=1}^C -\log  \frac{A^r_{jj}}{\sum_{i=1}^C A^r_{ij}},
\end{aligned}
\end{equation}
where $A^r$ is the attention matrix in the $r$th decoder layer. 

The objective function of learning the decoder is 
\begin{equation}
\label{eq:dec_loss}
    \Lagr_{dec} = \Lagr_{\tilde{\text{2D}}} + \Lagr_{\tilde{\text{3D}}} + \alpha\Lagr_{con},
\end{equation}
where $\alpha$ is a positive constant.

\subsection{Implementation Details}
\label{sec:implementation}

The proposed method is implemented in PyTorch. 
ResNet-50~\cite{he2016deep} pre-trained on ImageNet~\cite{russakovsky2015imagenet} serves as the 2D feature extractor while MinkowskiNet~\cite{choy20194d} works as the 3D feature extractor in the experiments. 
The numbers of heads, encoder layers, interlaced blocks, embedding dimension, and the width of FFN in the transformer are set to $4$, $3$, $2$, $96$, and $96$, respectively. 
$16$ multi-view images are randomly sampled for each scene.
The model is trained on eight NVIDIA 3090 GPUs with $500$ epochs.
The batch size, learning rate, and weight decay are set to $32$, $10^{-2}$, and $10^{-4}$, respectively. 
We use AdamW~\cite{kingma2014adam} as the optimizer. The weight $\alpha$ for $\Lagr_{con}$ is set to $0.5$.

\paragraph{Inference.}
Given a point cloud $P$ for inference, we feed it into our 3D encoder for feature extraction. 
The 3D CAM $\tilde{F}_\text{3D}^{s} \in \mathbb{R}^{S\times C}$, \ie, the segmentation result, is then obtained by passing the extracted features into the class-aware layer, as specified in Section~\ref{sec:encoder}. 
In MCTformer~\cite{xu2022multi}, 3D CAM can be further refined by the class-to-voxel attention maps $A_{c2s} \in \mathbb{R}^{C\times S}$ from the last $K$ transformer encoder layers, where $K=3$.
The refined 3D CAM is obtained through element-wise multiplication between CAM and the attention maps: $F = \tilde{F}_\text{3D}^{s} \odot A_{c2s}$, where $\odot$ denotes Hadamard product.
In addition, we consider the class-to-voxel attention maps in the interlaced decoder, if multi-view images are provided, which can be extracted from all the even layers, producing another refined 3D CAM $\hat{F}$.
Finally, the segmentation results can be obtained by applying the element-wise max operation to $F$ and $\hat{F}$.

We followed the common practice in ~\cite{Chang_2020_CVPR,xu2020weakly,liu2021one,wei2020multi,ren20213d,yang2022mil} to generate pseudo-segmentation labels by running inference on the training set. Then use a segmentation model, \eg Res U-Net~\cite{hou2021exploring}, to train on the pseudo labels with high confidence, \ie over $0.5$, and derive the segmentation model with $150$ epochs. No further post-processing is applied.

%% file: image/decoder.tex
\begin{figure}[t]
\begin{center}
{
\includegraphics[width=0.99\columnwidth]{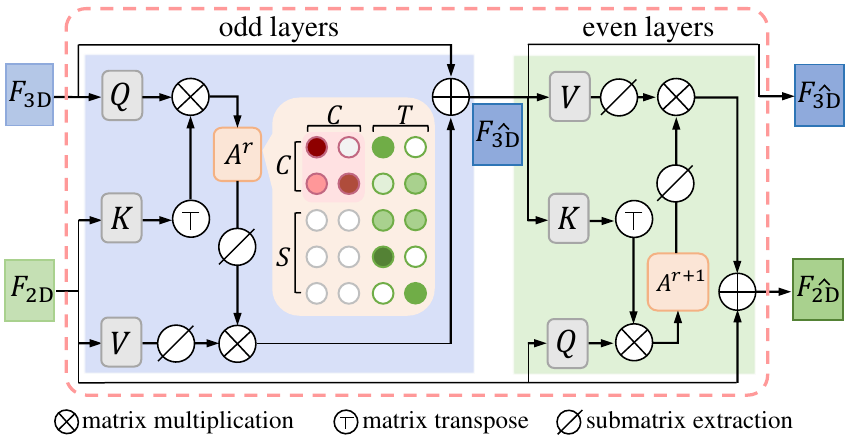}
}
\end{center}
\vspace{-0.15in}
\caption{
The architecture of an interlaced block. 
The multilayer perceptron with residual learning is not present for simplicity but is used in the block.
}
\label{fig:decoder}
\vspace{-0.1in}
\end{figure}

%% file: 4_experiment.tex
\section{Experimental Results}
\label{sec:experiments}

This section evaluates the proposed method. We begin by introducing the datasets and evaluation metrics. The competing methods are then presented and compared. Finally, we present ablation studies for each proposed component and analysis of our method.

\subsection{Datasets and Evaluation Metrics}
\label{sec:datasets}
The experiments were conducted using two large-scale point cloud datasets with multi-view images, S3DIS~\cite{armeni20163d} and ScanNet~\cite{dai2017scannet}.
S3DIS~\cite{armeni20163d} contains 272 scenes from six indoor areas. A total of 70,496 RGB images are collected. Each scene is represented by a point cloud with 3D coordinates and RGB values. Each point and pixel is labeled with one of 13 categories. Following previous works~\cite{qi2017pointnet, qi2017pointnet++,ren20213d, wang2019dynamic, xu2020weakly}, area 5 is taken as the test scene.
ScanNet~\cite{dai2017scannet} includes 1,201 training scenes, 312 validation scenes, and 100 test scenes with 20 classes. Over 2.5 million RGB images are collected. Following~\cite{hu2021bidirectional}, we sample one image out of every twenty to avoid redundancy in image selection.
The mean intersection over Union (mIoU) is employed as the evaluation metric for both datasets.

\subsection{Competing Methods and Comparisons}
We compare our MIT with competing weakly supervised point cloud segmentation and 2D-3D feature fusion methods.

\subsubsection{Point Cloud Segmentation Method Comparison}
We compare our proposed method to state-of-the-art methods for segmenting point clouds with scene-level supervision.
We also consider methods utilizing different supervision signals and extra data as input.
To begin with, we present the fully supervised methods~\cite{choy20194d,qi2017pointnet++,robert2022learning, wang2022semaffinet} for point cloud segmentation as they suggest potential performance upper bounds. Next, we show the methods~\cite{liu2021one,yu2022data,kweon2022joint} that employ various types of weak labels. Finally, we compare the segmentation methods~\cite{ren20213d, wei2020multi, yang2022mil} utilizing scene-level labels that indicate whether each class appears in the scene. 

Table~\ref{tab:sota} reports the mIoU results of the competing methods using different types of supervision or extra input data, such as RGB images, camera poses, or depth maps.
Existing methods that fuse 2D images with 3D data have demonstrated superior performance compared to 3D-only methods. However, the reliance on camera poses or depth maps limits their applicability. In contrast, our MIT can benefit from 2D images without such requirements, enhancing its generalizability.

\input{table/vs_sota}

By using efficient scene-level annotation, our MIT with 3D data only (the blue-shaded region in Figure~\ref{fig:model})
shows comparable results to the state-of-the-art weakly supervised method~\cite{ren20213d}, demonstrating the effectiveness of transformer encoder with the multi-class token~\cite{xu2022multi}.
The proposed interlace decoder further enhances the performance of the MIT with 3D-only data by incorporating the 2D image information.
Without introducing extra annotation costs, our method with 2D-3D fusion outperforms the existing methods by a large margin on both the ScanNet and S3DIS datasets.
This result demonstrates once again that 2D and 3D data are complementary. More importantly, the proposed method is capable of utilizing their complementarity in a weakly supervised manner.

Kweon~\etal~\cite{kweon2022joint} also confirms the effectiveness of combining 2D-3D data. However, their method requires non-negligible extra annotation costs on the images. According to ~\cite{bearman2016s, yang2022mil}, their method incurs more than five times the annotation cost required for scene-level annotation and even more than the sparsely labeled points setting.

We summarize the advantages of the scene-level setting in three aspects:   
\textbf{1)} \textbf{Efficiency}: Scene-level supervision is much more efficient to collect than other weak supervision types. 
According to~\cite{ren20213d,yang2022mil}, the labeling cost of sparse points ($1\%$ of points in ScanNet) is more than ten times higher than our scene-level setting.
\textbf{2)} \textbf{Generalization}: Our method based on scene-level supervision can be extended to other forms of weak supervision. Section~\ref{weak_cost} evaluates our method trained with diverse weak supervision types. 
\textbf{3)} \textbf{Potential}: Existing weakly supervised point cloud segmentation methods focus on the sparse-point supervision setting and achieve performances almost as good as fully supervised ones. Therefore, working with lower annotation costs, such as scene-level tags, shows potential and is worth exploring. Our method effectively carries it out by utilizing information from unlabeled images.

\subsubsection{2D-3D Fusion Method Comparison}
As far as we are aware, our MIT is the first attempt at exploring 2D-3D fusion without \textit{poses}, and the model is derived through \textit{scene-level supervision}. Hence, there is no existing fusion method for performance comparison.
To evaluate our method, we explore two approaches for 2D-3D feature fusion. First, we design a baseline method using a simple multi-layer perceptron for 2D-3D fusion. For each 3D voxel, we locate the nearest 2D pixel and concatenate the 3D feature with the 2D feature, followed by a 1×1 convolution to perform 2D-3D feature fusion. Second, we employ the bidirectional projection module~\cite{hu2021bidirectional} for 2D-3D fusion, which utilizes the pixel-to-point link matrix to fuse the 2D-3D features.

We apply the 2D-3D fusion methods on a weakly supervised point cloud segmentation method, MIL-transformer~\cite{yang2022mil}, as well as our proposed method.
Table~\ref{tab:fuse} provides the mIoU results of the competing 2D-3D fusion methods.
Our proposed interlaced decoder achieves superior results compared to the two competing 2D-3D fusion methods. More importantly, the interlaced decoder implements 2D-3D fusion without using poses or depths and performs even better when camera information is available (more details in Section~\ref{se4:extend_to_pose} and supplementary materials).
\input{table/fusion}

\input{image/scannet}

Our interlaced decoder offers the following advantages.
\textit{Multi-view aggregation}: The view quality differs in different views of the same 3D point, such as occlusion, or no 2D-3D correspondence.
Through the attention mechanism, the decoder learns how to effectively aggregates the multi-view information based on the semantic information.
\textit{Global attention}: The decoder can capture long-range dependencies, \ie, the receptive field is the whole scene. 
\textit{Low overhead}: The computational bottleneck of the decoder lies in cross-attention, whose complexity is linear to $N_\text{2D} \times N_\text{3D}$, where $N_\text{2D}$ and $N_\text{3D}$ are the numbers of 2D and 3D tokens, respectively.
Since we cast each 2D view into a token via global average pooling, $N_\text{2D} = C + T$, where $C$ is the number of classes and $T$ is the number of 2D views. As shown in Table~\ref{tab:n_view}, we can achieve good results by giving $T=16$ views; hence $N_\text{2D}$ can be small. 
To summarize, the proposed interlaced decoder introduces an acceptable cost but provides multi-view aggregation with global attention. Moreover, our interlaced decoder can further enrich features in 2D and 3D domains for better 3D segmentation.

\subsubsection{Qualitative Results}
Figure~\ref{fig:scannet} shows the qualitative results of our MIT with and without using the complementary 2D data.
By utilizing both 3D and 2D data, our method achieves promising segmentation results without using any point-level supervision.
With the help of detailed texture features in the 2D image, our method is able to classify objects with very similar geometric shapes, for example, door and wall. Take the second row of Figure~\ref{fig:scannet} as an example; MIT successfully segments the points belonging to the door by cooperating with the correct prediction from our 2D view (marked in blue), while the 3D only MIT fails to locate the points of the door, by only considering geometric and color features.

In addition, the category co-occurrence issue could hinder the optimization of the model with 3D data only.
Since optimization is based on scene-level labels, it is difficult to learn discriminative features for those co-occurring categories. 
As demonstrated in the second and third rows of Figure~\ref{fig:scannet}, MIT (3D-only) often fails to classify the chairs and tables since these categories often co-occur in a scene.
In contrast, our method leverages multi-view information during training. As each view only captures a small part of the scene, the issue of category co-occurrence could be alleviated, resulting in better segmentation performance.
With the proposed interlaced decoder, the network can learn more corresponding features between view and voxel under weak supervision. Additionally, the data tokens with position embedding and class tokens with contrastive loss facilitate the linking of views and voxels.

\subsection{Ablation Studies and Performance Analysis}
To evaluate the effectiveness of the proposed components, we perform ablation studies and analyze their performance.
We present ablation studies to evaluate the impacts of the proposed components and provide performance analysis. 

\subsubsection{Contributions of Components}
To evaluate the effectiveness of each proposed component, we first construct the baseline by considering only 3D data and utilizing class activation maps~\cite{wei2020multi, xu2020weakly}.
Then, we assess the contributions of each component, including the multi-class token transformer encoder (MCT), the interlaced decoder (Interlaced), and the N-pair loss ($\Lagr_{con}$), by successively adding each one to the baseline.
In addition, we evaluate the roles for 2D and 3D, as query and key-value pairs, by switching them. The result of the standard transformer decoder is also reported (the third row of Table~\ref{tab:ablation}) by taking 3D as query and 2D as key-value. Table~\ref{tab:ablation} illustrates the performance when using different combinations of the proposed modules and loss. The results validate that each component contributes to the performance of our method.

\input{table/ablation}

\subsubsection{Performance Analysis}
We discuss the extension of the proposed method and evaluate our method with different parameters and synthesized images in the following.
\paragraph{Extension with known poses and depths.}
\label{se4:extend_to_pose}
When camera poses, and depth maps are available, the correspondence between 3D world coordinates and 2D pixels can be established. 
Therefore, we can explicitly construct the position correlation between 2D views and 3D voxels. 
To this end, we first generate the 3D world coordinate maps for each view by following Yu~\etal~\cite{yu2022data}. 
All the 3D coordinate maps are fed into the coordinate embedding module $f_{emb}$ to obtain positional embedding, which is then added to the 2D image features. 
Through explicit positional information between the 2D view and 3D voxel, we can further boost performance, as shown in the last row of Table~\ref{tab:fuse}.

\paragraph{Analysis of parameters.}
We explore the influence of the number of 2D views and interlaced blocks by evaluating the quality of pseudo labels on the training set.
Table~\ref{tab:n_view} shows the performance of our MIT with different numbers of views used in the transformer. We found that performance is stable when given a sufficient number of views, as also reported in~\cite{hu2021bidirectional}.
Table~\ref{tab:n_block} presents the performance by altering the number of the proposed interlaced blocks. The results indicate that stacking two interlaced blocks performs the best while being saturated by adding more blocks.

\paragraph{Experiments with virtual image rendering.}
Among the limitations of our method is the need for multi-view 2D images within the 3D dataset. 
A potential solution would be the virtual view rendering of the 3D data. 
Several studies~\cite{kundu2020virtual,mccormac2017scenenet } suggest that synthesized images can further improve 3D segmentation performance. 
With the help of virtual view rendering ~\cite{kundu2020virtual}, our method still achieves competitive results ($34.3\%$ mIoU on ScanNet validation set) using the synthesized RGB images. 

\input{table/n_views}
\input{table/n_layers}

\paragraph{Extensions to other weak supervisions.}\label{weak_cost}
Thanks to the flexibility of the transformer, our method can be easily adapted to other weakly supervised settings, such as additional image-level labels, subcloud-level annotation, or sparsely labeled points. 

For the extra image-label annotation, it provides the class tag indicating the existing object category within each view image. Several methods~\cite{sun2020mining,ahn2018learning} are proposed to derive a 2D segmentation model based on this supervision and achieve promising results. Our method can easily train on image-level supervision by computing the multi-label classification loss on each image token.

Regarding the subcloud-level annotation, it sequentially crops a sphere point cloud from the scene and labels the existing objects within the sphere. This type of supervision alleviates the severe class imbalance issue in scene-level supervision. Our approach can be directly trained on subcloud-level supervision by considering the corresponding multi-view images in the subcloud.

For the setting with sparsely labeled points~\cite{liu2021one ,yang2022mil}, we can calculate the cross entropy loss on the self-attended voxel features $\hat{F}_\text{3D}$ and the labeled points. Furthermore, we note that the sparsely labeled 3D points can be projected onto the 2D image pixels, generating 2D pixel annotation. In spite of this, we do not explore this operation in our experiments and leave it for future research.

\input{table/cost}

Table~\ref{tab:anno_cost} shows the performance of our method under different types of weak supervision and the corresponding annotation cost. While scene-level annotation is the most efficient~\cite{yang2022mil}, its performance has room for improvement. The extra image-level labels can improve the performance of scene-level supervision but introduce additional burdens due to the large number of view images in each scene. According to ~\cite{bearman2016s}, the image-level labels cost $20$ seconds per image. In line with~\cite{kweon2022joint}, which utilized $17$ multi-view images in their setting, we used $16$ images per scene, resulting in an additional five-minute annotation time.
Even though both image-level and subcloud-level supervision types do not require point-level annotation, they could require more annotation efforts due to the large number of views and subclouds that need to be annotated. Sparsely labeled points, on the other hand, may perform better with less annotation effort.

Our approach can work effectively with diverse weak supervision, allowing for flexible savings in annotation costs. More importantly, our MIT shows promising results by using the most efficient scene-level supervision, while other weakly supervised methods cannot be straightforwardly applied in this scenario.

\input{table/backbone}
\paragraph{Experiments with different backbones.}
Table~\ref{tab:backbone} provides the performance (pseudo-label quality in mIoU) of our method on ScanNet with different 2D and 3D backbones, including different versions of 2D ResNet and 3D ResUNet. Our method's performance is consistent across different backbones.

%% file: table/vs_sota.tex
\begin{table}[!t]
	\centering
    \normalsize
	\resizebox{0.99\linewidth}{!}{
	\begin{tabular}{lccccccc}
		\toprule[1pt] 
		\multirow{2}{*}{Method}               &  \multirow{2}{*}{Sup.} &  \multicolumn{3}{c}{Extra inputs}   &   \multicolumn{2}{c}{ScanNet}  & {S3DIS}   \\
		 &   &  RGB & Pose   &  Depth   &      Val       &      Test   &   Test       \\ \midrule
		MinkUNet~\cite{choy20194d} & $\mathcal{F.}$ & - & -& - & 72.2 & 73.6 & 65.8       \\ 
		DeepViewAgg~\cite{robert2022learning}  & $\mathcal{F.}$  & \checkmark & \checkmark  & -  & 71.0 & - & 67.2     \\      
		SemAffiNet~\cite{wang2022semaffinet}  & $\mathcal{F.}$  & \checkmark & \checkmark  & \checkmark  & - & 74.9 & 71.6     \\   
        \midrule
            OTOC~\cite{liu2021one} & $\mathcal{P.}$ & - & - & -&  - & 59.4 & 50.1 \\
            Yu \etal~\cite{yu2022data} & $\mathcal{P.}$ & \checkmark  & \checkmark  &  \checkmark &  - & 63.9 & -     \\ 
        \midrule
            Kweon \etal~\cite{kweon2022joint} & $\mathcal{S.+I.}$ & \checkmark & \checkmark & -& 49.6 & 47.4 & -     \\ 
        \midrule
		MPRM~\cite{wei2020multi} & $\mathcal{S.}$  & - & -& -&  24.4 & - & 10.3      \\
		MIL-Trans~\cite{yang2022mil} & $\mathcal{S.}$& - & -& -&  26.2 & - & 12.9      \\
		WYPR~\cite{ren20213d}& $\mathcal{S.}$ & - & -& -&  29.6 & 24.0 & 22.3      \\
		MIT (3D-only) & $\mathcal{S.}$ & - & -& -&  31.6 & 26.4 & 23.1 \\ 
		\textbf{MIT (Ours)} & $\mathcal{S.}$ & \checkmark & -& -& \textbf{35.8} & \textbf{31.7} & \textbf{27.7} \\ 
		\bottomrule[1pt]

	\end{tabular}}
	\caption{
Quantitative results (mIoU) of several point-cloud segmentation methods with diverse supervisions and input data settings on the ScanNet and S3DIS datasets.
``Sup.'' indicates the type of supervision. ``$\mathcal{F.}$'' represents full annotation. ``$\mathcal{P.}$'' gives sparsely labeled points. ``$\mathcal{S.}$'' denotes scene-level annotation. ``$\mathcal{I.}$'' implies image-level annotation.
	}
	\label{tab:sota}
\end{table}

%% file: table/fusion.tex
\begin{table}[!t]
\small
\centering
	\resizebox{0.8\linewidth}{!}{
	\begin{tabular}{lcccc}
		\toprule[1pt] 
		Method     & Fusion  & Pose        &  Depth     & mIoU   \\ \midrule
        MIL-Trans~\cite{yang2022mil} & MLP &  $\checkmark$ &   - & 25.6  \\
        MIL-Trans~\cite{yang2022mil} & BPM~\cite{hu2021bidirectional} &  $\checkmark$ &   $\checkmark$ & 25.9  \\ 
        MIT         &  MLP      &  $\checkmark$     &  - & 32.6        \\ 
        MIT        &  BPM~\cite{hu2021bidirectional}      &  $\checkmark$     &  $\checkmark$ & 32.4        \\ 
        MIT       &  Interlaced      &  -   &  -  & 35.8       \\ 
        MIT       &  Interlaced      &  $\checkmark$   &  $\checkmark$  & \textbf{37.1}       \\ 
		\bottomrule[1pt]
		\vspace{-0.1in}
	\end{tabular}}

\caption{Quantitative results (mIoU) of our method (interlaced decoder) and competing methods with different 2D-3D fusion strategies on the ScanNet validation set using scene-level annotations.}
\label{tab:fuse}
\end{table}

%% file: image/scannet.tex
\begin{figure*}[t]
    \centering
    \includegraphics[width=.95\linewidth,keepaspectratio]{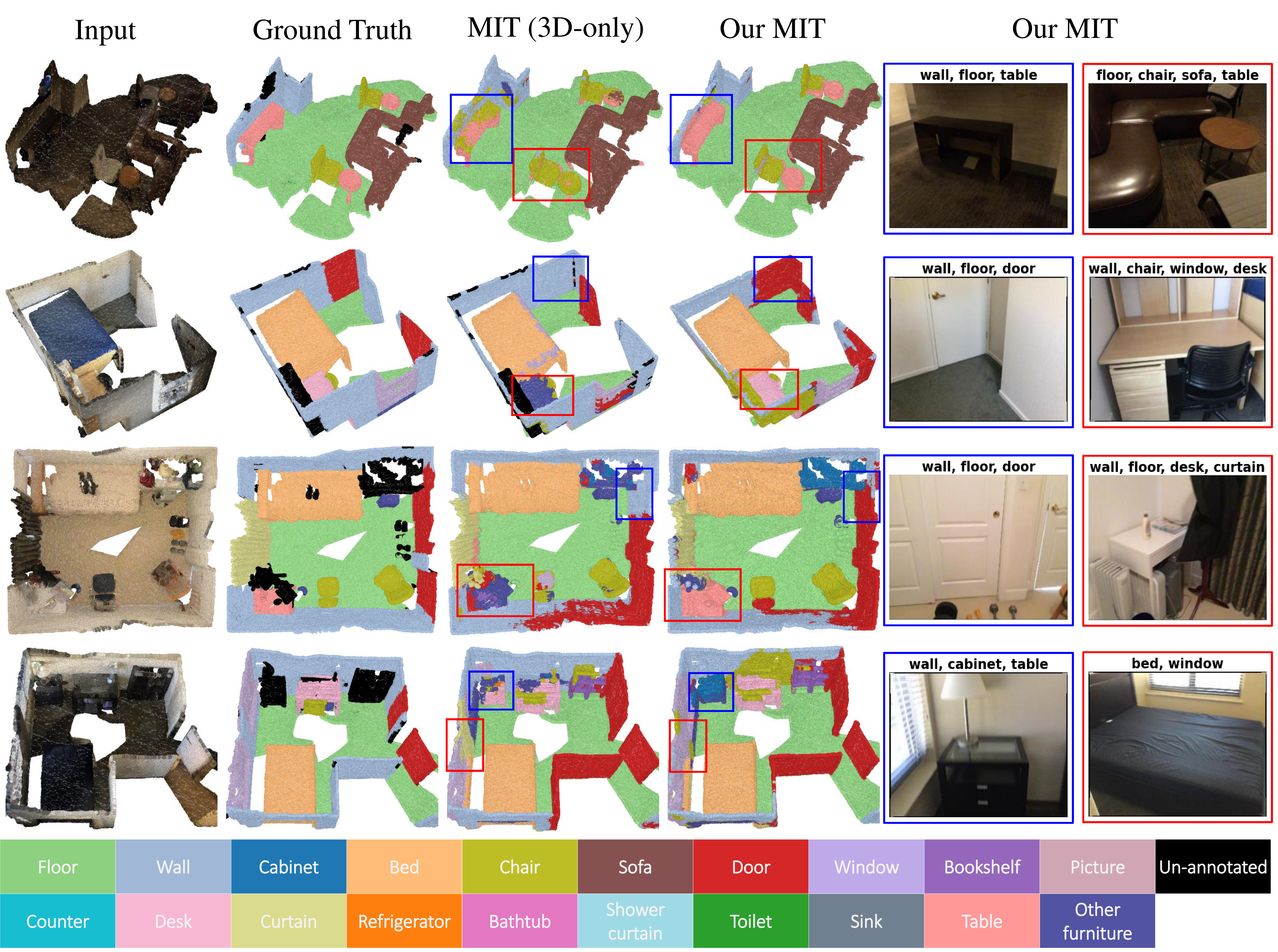}
    \vspace{-.25cm}

    \caption{
    Qualitative results on the ScanNet dataset with scene-level supervision. The colored boxes highlight the differences between our MIT and MIT with 3D data only, and their corresponding views are shown on the right with outlines of the same color. 
    For each view, the tags at the top indicate the results of the multi-label classification.
    }
    \label{fig:scannet}
\end{figure*}

%% file: table/ablation.tex
\begin{table}[!t]
	\centering
    \scalebox{0.9}{
	\begin{tabular}{cc|ccc|cc}
		\toprule
		Query        &  Key-Value   &  MCT  & Interlaced & $\Lagr_{con}$ & mIoU  \\ \midrule
         -    &  -          &          &  &              &   26.1     \\ 
         -    &  -           & $\checkmark$ &         &          &   31.6      \\ 
        \midrule
        3D    & 2D           & $\checkmark$ &                    &    &     33.7     \\
        3D    & 2D           & $\checkmark$ & $\checkmark$       &    &     35.4       \\
        2D    & 3D           & $\checkmark$ & $\checkmark$       &    &     35.2       \\
        3D    & 2D           & $\checkmark$ & $\checkmark$       &   $\checkmark$         & \textbf{35.8}  \\
    	\bottomrule
	\end{tabular}}
		\caption{
		The mIoU performance of different combinations of proposed components on the validation set of the ScanNet dataset. ``Query'' and ``Key-Value'' denote the input to the decoder. ``MCT'' and ``Interlaced'' are the multi-class tokens encoder and decoder architectures respectively. $\Lagr_{con}$ denotes the contrastive loss on the class tokens.
	}
	\label{tab:ablation}
\end{table}

%% file: table/n_views.tex
\begin{table}[!t]
	\centering
	\renewcommand{\tabcolsep}{6.5pt}
	\resizebox{0.8\linewidth}{!}{
	\begin{tabular}{ccccc}
		\toprule[1pt] 
		Number of views & 4   & 16   &   32 & 64   \\ \midrule
		mIoU         &  29.7   & 32.7 &   30.9 & 31.2 \\
		\bottomrule[1pt]
	\end{tabular}}
		\caption{
		Performance with different numbers of views on the mIoU of pseudo labels on the ScanNet.
	}
	\label{tab:n_view}
\end{table}

%% file: table/n_layers.tex
\begin{table}[!t]
	\centering
	\renewcommand{\tabcolsep}{6.5pt}
	\resizebox{0.85\linewidth}{!}{
	\begin{tabular}{ccccc}
		\toprule[1pt] 
		$R$ interlaced blocks &      1  & 2   &     3  &     4 \\ \midrule
		mIoU     & 31.4 & 32.7 &     32.1&     32.4\\
		\bottomrule[1pt]
	\end{tabular}}
		\caption{
		Performance with different numbers of interlaced blocks on the mIoU of pseudo labels on the ScanNet.
	}
	\label{tab:n_block}
\end{table}

%% file: table/cost.tex
\begin{table}[tbp]
  \small
  \centering
     \resizebox{0.999\linewidth}{!}{
    \begin{tabular}{ccccc}
    \toprule
             & Scene & Scene+Image & Subcloud & 20pts  \\
     \midrule
     mIoU   & 35.8 & 45.4 & 46.8 & 61.9       \\ 
     label effort & $<$ 1 min & 5 min  & 3 min    & 2 min     \\ \bottomrule
    \end{tabular}%
    }
  \caption{
  The mIoU performance of our MIT and its average annotation time per scene of different weak supervisions on ScanNet.
  }    
  \label{tab:anno_cost}%
\end{table}%

%% file: table/backbone.tex
\begin{table}[t]
\small
\centering
	\resizebox{0.65\linewidth}{!}{
	\begin{tabular}{ccc}
		\toprule[1pt] 
		3D Backbone     & 2D Backbone       & mIoU   \\ \midrule
        ResUNet-18 & ResNet-50 & 32.7   \\
        ResUNet-18 & ResNet-101  & 33.1  \\ 
        ResUNet-34  &  ResNet-101       & 32.9      \\ 
		\bottomrule[1pt]
		\vspace{-0.1in}
	\end{tabular}}

\caption{Performance with different backbones on the mIoU of pseudo labels on the ScanNet.}
\label{tab:backbone}
\end{table}

%% file: 5_conclusion.tex
\section{Conclusion}
\label{sec:conclusion}

This paper presents a multimodal interlaced transformer, MIT, for weakly supervised point cloud segmentation.
Our method represents the first attempt at 2D and 3D information fusion with scene-level annotation.
Through the use of the proposed interlaced decoder, which performs implicit 2D-3D feature fusion via cross-attention, we are able to effectively fuse 2D-3D features without using camera poses or depth maps. 
Our MIT achieves promising performance without using any point-level or pixel-level annotations.
Furthermore, we develop class token consistency to align the multimodal features. 
MIT is end-to-end trainable. It has been extensively evaluated on two challenging real-world large-scale datasets.
Experiments show that our method performs favorably against existing weakly supervised methods.
We believe MIT has the potential to enhance other recognition tasks that involve both 2D and 3D observations, in an efficient manner.

\vspace{-0.1cm}
\paragraph{Discussion and future work.}
Our current method has not utilized the spatial information conveyed in the images since global average pooling is applied to the image features. 
We attempted to flatten image features instead of using global average pooling to obtain patch tokens, similar to~\cite{liu2022petr,dosovitskiy2020image}, but achieved inferior results. One possible reason is that a large number of patch tokens introduces noise under scene-level supervision. A solution to this issue can achieve joint 2D-3D segmentation with weak supervision, which could be an interesting area for future research.

\vspace{-0.1cm}
\paragraph{Acknowledgements.}
This work was supported in part by the National Science and Technology Council (NSTC) under grants 111-2628-E-A49-025-MY3, 112-2221-E-A49-090-MY3, 111-2634-F-002-023, 111-2634-F-006-012, 110-2221-E-002-124-MY3, and 111-2634-F-002-022. This work was funded in part by MediaTek, Qualcomm, NVIDIA, and NTU-112L900902.

%% file: 6_supplementary.tex
\section{Supplementary}

This document provides additional experiments. In the following, we first present some potential extensions of the proposed method. Then, we provide the model architecture and running time. Lastly, we present detailed quantitative results as well as qualitative examples, including less successful cases.

\subsection{Extensions of the Proposed Method}

\subsubsection{Extension with Known Poses and Depths}
In Section~\ref{se4:extend_to_pose}, we present how to extend our method when camera poses and depth maps are available. In the following, we elaborate on this extension by providing further details.

Inspired by the fact that positional information can enrich image features~\cite{liu2022petr}, we perform positional embedding for both 2D and 3D features before passing them to the transformer encoders. 
This way, both 2D and 3D share a common 3D world space, facilitating explicit position correlation between 2D images and 3D point clouds.
We first generate the 3D coordinate map $\mathbf{x}_t \in \mathbb{R}^{H \times W\times 3}$ for each view $\mathbf{v}_t$. 
Given the depth map $\mathbf{d}_t$ and camera projection matrix $\mathbf{k}_t$, the 3D world coordinate $\mathbf{x}_t(u,v)$ at 2D position $[u, v]$ is computed by
\begin{equation}
[(\mathbf{x}_t(u,v))^{\top}, 1]^{\top} = \mathbf{d}_t(u,v) \cdot \mathbf{k}_t^{-1}[u,v,1]^{\top}.
\label{eq:3D}
\end{equation}

Via Eq.~\ref{eq:3D}, we obtain the 3D world coordinate map $\mathbf{x}_t$ for each view image $\mathbf{v}_t$. 
All $T$ 3D coordinate maps $\{\mathbf{x}_t\}_{t=1}^{T}$ are fed into a coordinate embedding module $f_{emb}$, which is composed of two $1 \times 1$ convolution layers with ReLU activation, to get the positional embedding $z_\text{2D} \in \mathbb{R}^{T \times H \times W \times D }$, where $D$ is the embedding dimension. 
The positional embedding is added to 2D features for 3D positional awareness.
Since each point of a point cloud $P$ lies in the 3D space, $f_{emb}$ is directly applied to all points and gets $z_\text{3D} \in \mathbb{R}^{M \times D}$, where $M$ is the number of points in $P$.
Figure~\ref{fig:model} depicts the extended method by revising Figure 2 in the submitted paper.

\input{image/model_supp}

\subsubsection{Extension to Joint 2D-3D Segmentation}

As discussed in Section~\ref{sec:conclusion}, the proposed method can be extended to joint 2D and 3D segmentation using weak supervision.
Through flattening the image features $F_\text{2D}$ instead of applying global average pooling, we obtain a set of multi-view patch tokens, which can be further considered as segmentation results~\cite{xu2022multi}. 
However, we get inferior results, $0.275$ and $0.129$ in mIoU of 3D and 2D, respectively, on the ScanNet training set. It may be because too many views generate many patch tokens and create lots of noise. Specifically, the image size is $320 \times 240$ in ScanNet, creating $80$ patch tokens for a view and a total of $1280$ tokens for $16$ views. The high number of 2D tokens hinders the optimization of self-attention and cross-attention, leading to unsatisfactory performance.

\subsection{Model Architecture and Running time}
As mentioned in Section~\ref{sec:implementation}, ResNet-50~\cite{he2015delving} is adopted as the 2D feature extractor. MinkowskiNet~\cite{choy20194d}  work as the 3D feature extractor for ScanNet and S3DIS. Specifically, we use MinkowskiUNet18A, and the voxel size is set to 5cm. The network was optimized on a machine with eight NVIDIA GTX 3090 GPUs. With 500 epochs, it took about two days to complete the optimization.
For the semantic segmentation model, we use MinkowskiUNet18C with the voxel size set to 2cm. The network was optimized on a machine with eight NVIDIA GTX 1080Ti GPUs. The optimization required 150 epochs, which took around a day to complete.

As discussed in Sec 4.2.2, the overhead of the proposed interlaced decoder is acceptable.
Table~\ref{tab:speed} shows the inference time and computational cost of different methods. 
WYPR~\cite{ren20213d} is not presented since their model code are not  \href{https://github.com/facebookresearch/WyPR/issues/10}{publicly available.}

\input{table/speed}

\subsubsection{Generalize to Another Backbone}
To assess the generalizability of our approach, we employ PointNet++~\cite{qi2017pointnet++} as the 3D feature extractor, which also serves as a popular backbone for point cloud-based applications. Notably, our method yields competitive performance results, achieving $35.1$ and $29.7$ mIoU on the ScanNet validation set and S3DIS test set, respectively. The results demonstrate that our MIT is general because it can work with different backbones.

\subsubsection{Utilizing no pre-trained 2D model}
We train our method with randomly initialized 2D ResNet-50 and observe only a minor performance drop ($35.8\%\!\rightarrow\!34.6\%$ in mIoU on the ScanNet validation set). This result indicates that our method does not rely heavily on ImageNet pre-training and can work with pure 3D scene-level supervision.

\subsection{Quantitative Results of Multi-view Images}
We also report the performance of multi-label image classification in Table~\ref{tab:multilabel}. This task aims to find all existing categories in a single view, by giving only the class appearance in the multi-view images for training.
We first report the baseline result, which is conducted by averaging the estimated class scores across all views during training and obtaining the per-view classification result by passing a single view to the model. ResNet-50~\cite{he2016deep} is adopted as the feature extractor.
Our method enriches the views with self-attention in 2D and cross-attention in 3D, showing better classification results compared to competing methods that only consider 2D information.

\input{table/multi_label}

\subsection{Class-wise Quantitative Performance}
In Table 1 of the submitted paper, we report the average
performance over the categories of ScanNet and S3DIS. Here we provide detailed class-wise results. Table~\ref{table:scannet_per_cls}, Table~\ref{table:scannet_test} and Table~\ref{table:s3dis_per_cls} show the class-wise performance of our
method on the ScanNet validation set, ScanNet test set, and S3DIS datasets, respectively.

\subsection{Qualitative Results}
Figure~\ref{fig:scan} and Figure~\ref{fig:s3dis} show the segmentation results generated by our method with scene-level supervision, including both successful cases and less successful ones. It can be observed that the proposed method delineates precise segmentation contours without using any point-level supervision. 
However, those categories with very similar shapes and colors lead to wrong segmentation results, such as the other furniture in the example of the last column of the third row in Figure~\ref{fig:scan} and the clutter in the example of the last column of the first row in Figure~\ref{fig:s3dis}. Also, some points of wall examples may be misclassified as doors or windows since they share very similar shapes.

\input{table/scannet}
\input{table/s3dis}
\input{image/scan}
\input{image/s3dis}

%% file: image/model_supp.tex
\begin{figure*}[t]
    \centering
    \includegraphics[width=.95\linewidth]{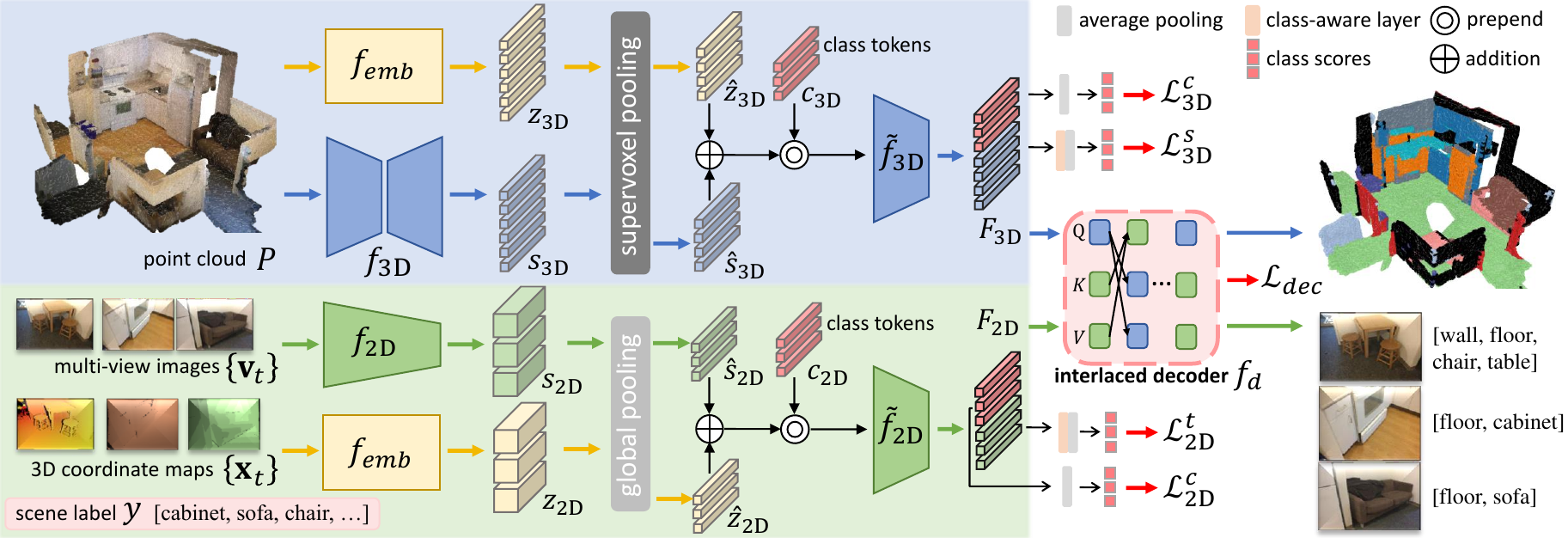}
    \vspace{-0.05in}
    \caption{
    Network architecture of our MIT extension with camera poses and depths maps.
    }
    \label{fig:model_supp}
    \vspace{-0.05in}
\end{figure*}

%% file: table/speed.tex
\begin{table}[!t]
  \small
  \centering
     \resizebox{0.75\linewidth}{!}{
           \begin{tabular}{ ccc }
              \toprule
                Methods       & Time &  FLOPs \\
              \midrule
          MIL-Trans~\cite{yang2022mil}   & 8.9 ms       & 181 G \\
        MIT (3D-only)   & 9.4 ms       & 199 G \\
        MIT (Ours)     & 27.3 ms       & 220 G  \\ 
              \bottomrule
            \end{tabular}
        
    }

  \caption{
  The inference time and FLOPs of different methods.
  }    
  \label{tab:speed}%
\end{table}%

%% file: table/multi_label.tex
\begin{table}[!t]
  \scriptsize
  \centering

     \resizebox{0.9\linewidth}{!}{
      \begin{tabular}{lccc}
        \toprule
        Method & Sup. & ScanNet & S3DIS \\ \midrule
          Baseline        &  $\mathcal{F.}$ &  79.4  & 82.3 \\ 
          Baseline        &  $\mathcal{S.}$  & 52.4  & 55.1\\ 
          Kim~\etal\cite{kim2022large} &  $\mathcal{S.}$ &  54.9  &58.2\\
          MIT (Ours)        &    $\mathcal{S.}$  &  56.1  &57.9\\
        \bottomrule
        \end{tabular}%
        
    }

  \caption{
  Quantitative results (mAP) of several multi-label classification methods with diverse supervision settings on the ScanNet and S3DIS image datasets. ``Sup.'' denotes the type of supervision. ``$\mathcal{F.}$'' represents full annotation for each view. ``$\mathcal{S.}$'' indicates that class tag annotation is shared by all views in the scene.
  }    
  \label{tab:multilabel}%
\end{table}%

%% file: table/scannet.tex
\begin{table*}[!ht]
    \footnotesize
    \centering 
    \scalebox{1.0}{       
    \setlength{\tabcolsep}{1pt}
    \begin{tabular}{l | c c c c c c c c c c c c c c c c c c c c | c}
        \toprule
        Method & wall & floor & cabinet & bed & chair & sofa & table & door & window & B.S. & picture & cnt & desk & curtain & fridge & S.C. & toilet & sink & bathtub & other & mIOU\\
        
        \midrule
        MIL-trans~\cite{yang2022mil} & 52.1 & 50.6 & 8.3 & 46.3 & 27.9 & 39.7 & 20.9 & 15.8 & 26.8 & 40.2 & 8.1 & 21.1 & 22.0 & 45.9 & 4.5 & 16.6 & 15.2 & 32.4 & 21.2 & 8.0 & 26.2\\   
        
        WYPR~\cite{ren20213d} & 52.0 & 77.1 & 6.6 & 54.3 & 35.2 & 40.9 & 29.6 & 9.3 & 28.7 & 33.3 & 4.8 & 26.6 & 27.9 & 69.4 & 8.1 & 27.9 & 24.1 & 25.4 & 32.3 & 8.7 & 31.1 \\        
        
        \textbf{MIT (Ours)} & 57.3 & 89.7 & 24.1 & 54.9 & 31.5 & 62.8 & 42.5 & 19.8 & 27.4 & 45.1 & 1.1 & 31.4 & 41.7 & 41.4 & 17.6 & 25.0 & 34.5 & 8.3 & 44.4 & 15.6 & 35.8 \\

        \bottomrule
    \end{tabular}
    }
\caption{Quantitative results (mIoU) of several point-cloud segmentation methods with scene-level supervision setting on the ScanNet validation set. ``B.S.'' denotes bookshelf; ``S.C.'' stands for shower curtain and ``cnt'' denotes counter. }  
\label{table:scannet_per_cls}
\end{table*}

\begin{table*}[!ht]
    \footnotesize
    \centering 
    \scalebox{1.05}{       
    \setlength{\tabcolsep}{1pt}
    \begin{tabular}{l | c c c c c c c c c c c c c c c c c c c c | c}
        \toprule
        Method & wall & floor & cabinet & bed & chair & sofa & table & door & window & B.S. & picture & cnt & desk & curtain & fridge & S.C. & toilet & sink & bathtub & other & mIOU\\
\midrule
        \textbf{MIT (Ours)} & 42.2 & 82.1 & 16.3 & 55.8 & 30.6 & 57.6 & 35.9 & 19.3 & 27.0 & 39.0 & 1.4 & 25.3 & 27.7 & 31.3 & 21.3 & 17.8 & 47.8 & 7.9 & 29.8 & 18.8 & 31.7 \\

        \bottomrule
    \end{tabular}
    }
\caption{Quantitative results (mIoU) of our method with scene-level supervision setting on the test set from official ScanNet benchmark server. ``B.S.'' denotes bookshelf; ``S.C.'' stands for shower curtain and ``cnt'' denotes counter. }  
\label{table:scannet_test}
\end{table*}

%% file: table/s3dis.tex
\begin{table*}[!ht]
    \footnotesize
    \centering 
    \scalebox{1.15}{   
    \setlength{\tabcolsep}{1pt}
    \begin{tabular}{l | c c c c c c c c c c c c c | c}
        \toprule
        Supervision& ceil & floor & wall & beam & column & window & door & chair & table & bookcase & sofa & board & clutter & mIOU\\
        
        \midrule
        MIL-Trans~\cite{yang2022mil} & 24.9 & 4.7 & 40.0 & 0.0 & 1.3 & 2.2 & 1.8 & 5.6 & 16.8 & 33.0 & 32.1 & 0.1 & 5.8 & 12.9 \\
        
        \textbf{MIT (Ours)} & 80.8 & 81.0 & 81.8 & 0.0 & 0.9 & 0.2 & 27.6 & 26.7 & 19.5 & 15.5 & 16.8 & 0.0 & 9.9 & 27.7\\
        \bottomrule
    \end{tabular}%
    }
\caption{Quantitative results (mIoU) of the proposed method with diverse supervision settings on the S3DIS Area 5 dataset.}  
\label{table:s3dis_per_cls}
\end{table*}

%% file: image/scan.tex
\begin{figure*}[t]
\centering

\includegraphics[width=0.95\textwidth]{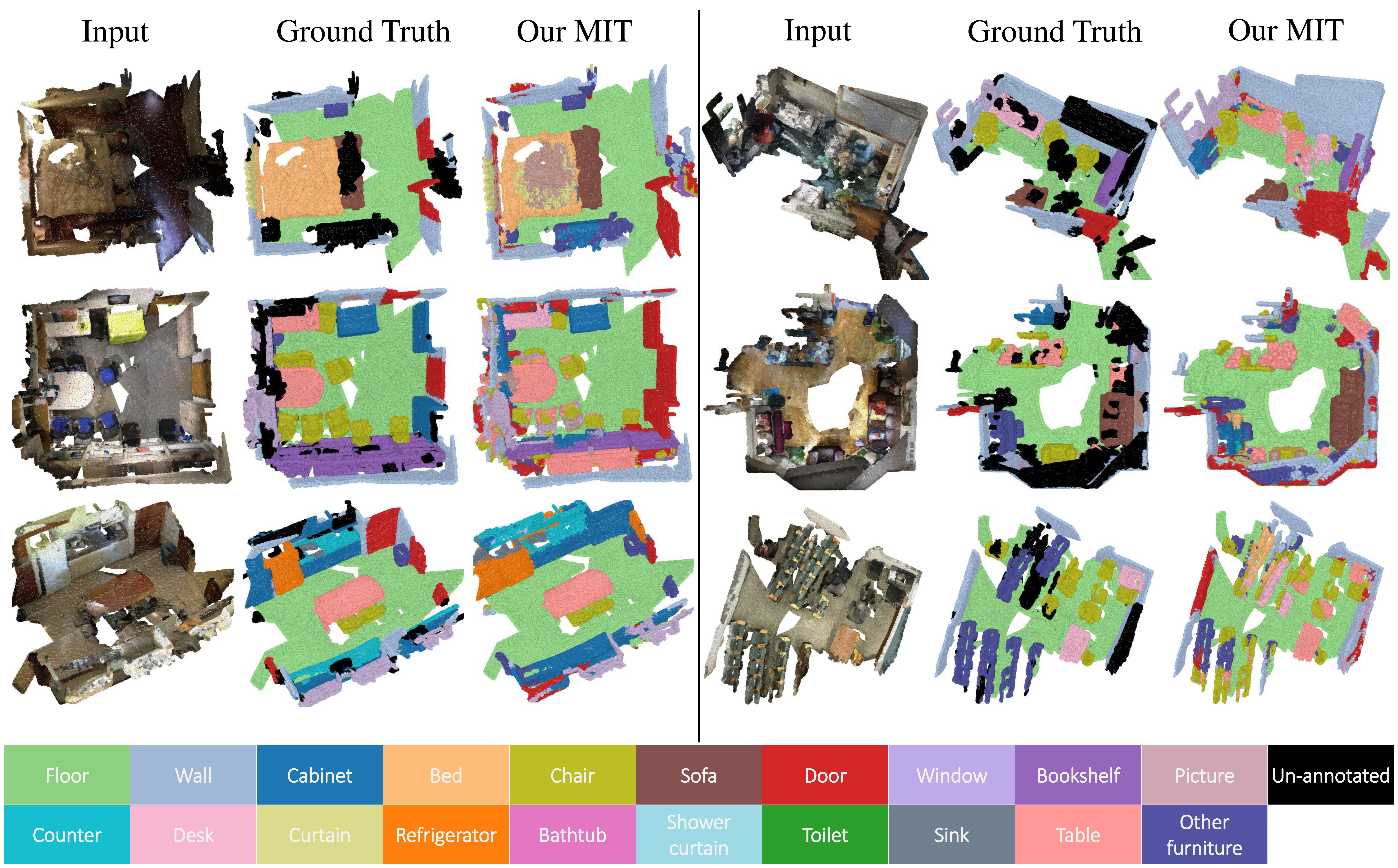}

\caption{
Qualitative results on the ScanNet dataset with scene-level supervision. Each category is associated with the same row. For each example, we show the input cloud, the ground-truth label, and our segmentation result. The last example of each row (on the right of the gray line) shows a less successful case.
}
\label{fig:scan}
\end{figure*}

%% file: image/s3dis.tex
\begin{figure*}[tp]
\begin{center}
{%
\includegraphics[width=0.99\textwidth]{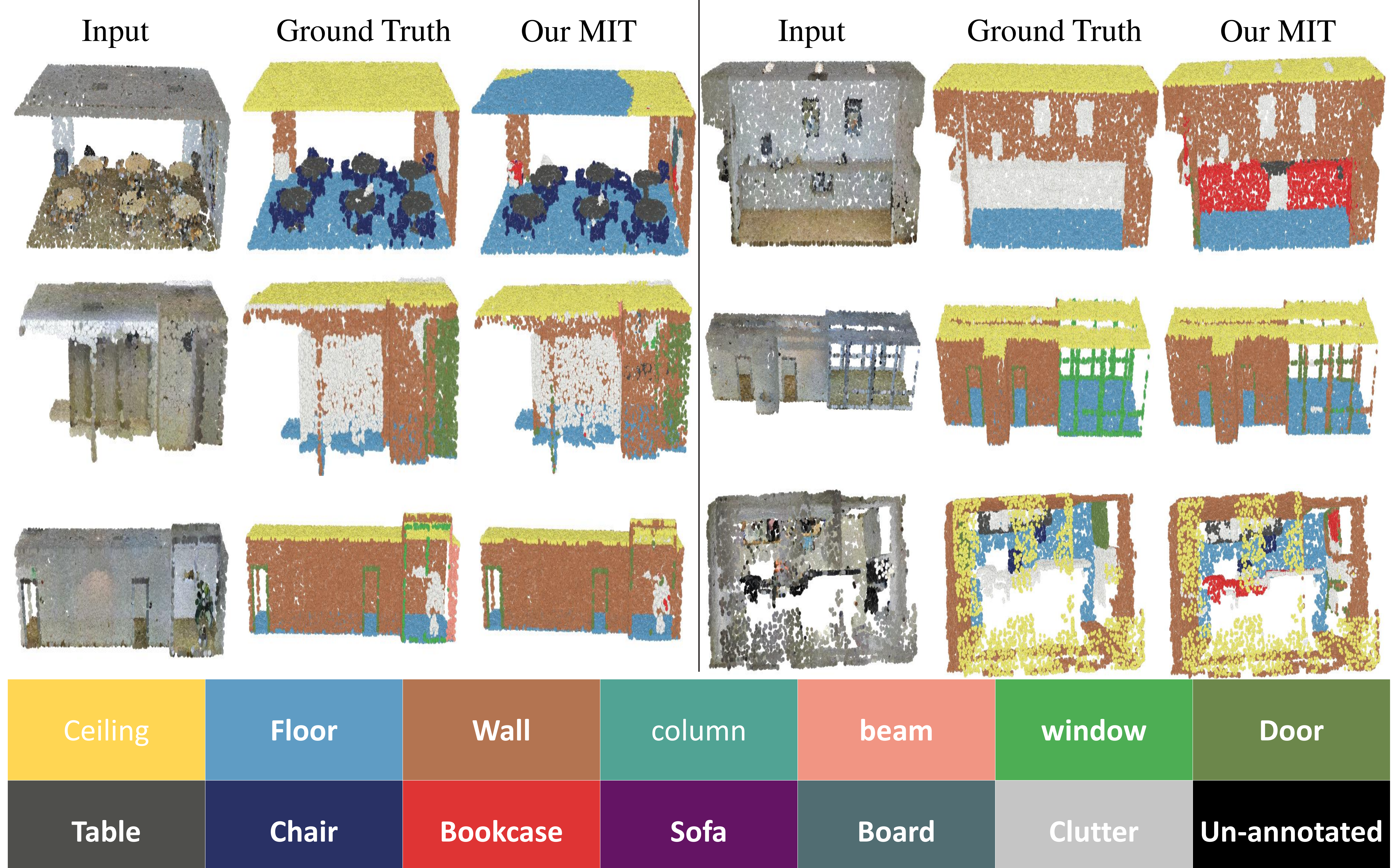}
}
\end{center}
\caption{
Qualitative results on the S3DIS dataset with scene-level supervision. Each category is associated with the same row. For each example, we show the input cloud, the ground-truth label, and our segmentation result. The last example of each row (on the right of the gray line) shows a less successful case.
}
\label{fig:s3dis}
\end{figure*}